\documentclass{article}
\usepackage{spconf,amsmath,graphicx,multirow}

\usepackage{times}

\title{The Microsoft 2017 Conversational Speech Recognition System}

\name{W. Xiong, L. Wu, F. Alleva, J. Droppo, X. Huang, A. Stolcke}
\address{Microsoft AI and Research\\
Technical Report MSR-TR-2017-39\\
August 2017}

\begin{document}

\sloppy

\maketitle

\begin{abstract}
We describe the 2017 version of Microsoft's conversational speech
recognition system, in which we update our 2016 system with recent
developments in neural-network-based acoustic and language modeling to
further advance the state of the art on the Switchboard speech recognition
task.
The system adds a CNN-BLSTM acoustic model to the set of model architectures 
we combined previously, and includes character-based and dialog session aware LSTM language
models in rescoring.
For system combination we adopt a two-stage approach, whereby subsets of 
acoustic models are first combined at the senone/frame level, followed by a word-level voting 
via confusion networks.
We also added a confusion network rescoring step after system combination.
The resulting system yields a 5.1\% word error rate on the 2000 Switchboard
evaluation set.
\end{abstract}

\section{Introduction}
	\label{sec:intro}

We have witnessed steady progress in the improvement of automatic speech recognition (ASR)
systems for conversational speech, a genre that was once considered among the hardest in the speech recognition community due
to its unconstrained nature and intrinsic variability \cite{GreenbergEtAl:icslp96}.
The combination of deep networks and efficient training methods with older neural modeling concepts
\cite{pineda1987generalization,williams1989learning,waibel1989phoneme,lecun1995convolutional,lecun1989backpropagation,robinson1991recurrent,hochreiter1997long}
have produced steady advances in both acoustic modeling
\cite{seide2011conversational,sak2014long,sak2015fast,saon2015ibm,sercu2016very,bi2015very,qian2016very}
and language modeling
\cite{mikolov2010recurrent,mikolov2012context,sundermeyer2012lstm,medennikov2016improving}.
These systems typically 
use deep convolutional neural network (CNN) architectures in acoustic modeling, and 
multi-layered recurrent networks with gated memory (long-short-term memory, LSTM \cite{hochreiter1997long}) models for both 
acoustic and language modeling, driving the word error rate on the benchmark Switchboard corpus \cite{godfrey1992switchboard}
down from its mid-2000s plateau of around 15\% to well below 10\%.
We can attribute this progress to the neural models' ability to learn regularities over a wide acoustic context in both time and frequency dimensions,
and, in the case of language models, to condition on unlimited histories and learn representations of functional word similarity
\cite{BengioEtAl:2006,mikolov2013linguistic}.

Given these developments, we carried out an experiment last year, to measure the accuracy of a state-of-the-art 
conversational speech recognition system against that of professional transcribers.
We were trying to answer the 
question whether machines had effectively caught up with humans in this, originally very challenging,
speech recognition task.
To measure human error on this task, we submitted the Switchboard evaluation data to our standard conversational speech 
transcription vendor pipeline (who was left blind to the experiment), postprocessed the output to remove text normalization discrepancies,
and then applied the NIST scoring protocol.
The resulting human word error was 5.9\%, not statistically different from the 5.8\% error rate achieved by our ASR system
\cite{parity-techreport}.
In a follow-up study \cite{StolckeDroppo:interspeech2017}, we found that qualitatively, too, the human and machine transcriptions were 
remarkably similar: the same short function words account for most of the errors, the same speakers tend to be easy or hard to transcribe, and 
it is difficult for human subjects to tell whether an errorful transcript was produced by a human or ASR.
Meanwhile, another research group carried out their own measurement of human transcription error \cite{ibm-human:interspeech2017},
while multiple groups reported further improvements in ASR performance \cite{ibm-human:interspeech2017,capio:interspeech2017}.
The IBM/Appen human transcription study employed a more involved transcription process with more listening passes, a pool of transcribers, and access to the conversational
context of each utterance, yielding a human error rate of 5.1\%.
Together with a prior study by LDC \cite{glenn2010transcription}, we can conclude that human performance, unsurprisingly, falls within a range depending
on the level of effort expended.

In this paper we describe a new iteration in the development of our system, pushing well past the 5.9\% benchmark we measured previously.
The overall gain comes from a combination of smaller improvements in all components of the recognition system.
We added an additional acoustic model architecture, a CNN-BLSTM, to our system.
Language modeling was improved with an additional utterance-level LSTM based on characters instead of words,
as well as a dialog session-based LSTM that uses the entire preceding conversation as history.
Our system combination approach was refined by combining predictions from multiple acoustic models at both the senone/frame and word levels.
Finally, we added an LM rescoring step after confusion network creation, bringing us to an overall error rate of 5.1\%,
thus surpassing the human accuracy level we had measured previously.
The remainder of the paper describes each of these enhancements in turn, followed by overall results.

\section{Acoustic Models}
	\label{sec:acoustic}

\subsection{Convolutional neural nets}

We used two types of CNN model architectures: ResNet and LACE
(VGG, a third architecture used in our previous system, was dropped).
The residual-network (ResNet) architecture \cite{he2015deep} is a standard CNN with added
highway connections \cite{DBLP:journals/corr/SrivastavaGS15}, i.e., a linear transform of each layer's input to the layer's output
\cite{DBLP:journals/corr/SrivastavaGS15,ghahremani2016linearly}.
We apply batch normalization \cite{batchnorm} before computing rectified linear unit (ReLU) activations.

\begin{figure}[t]
\centering
\includegraphics[width=0.45\textwidth]{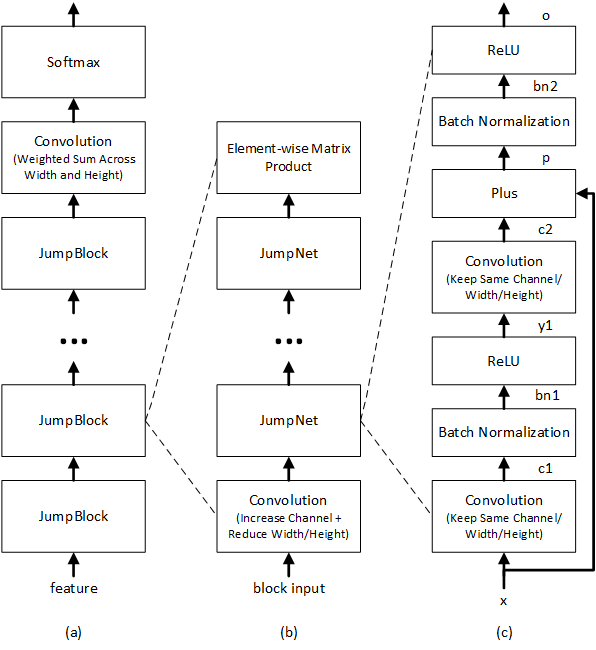}
\caption{LACE network architecture}
\label{fig:LACE}
\end{figure}

The LACE (layer-wise context expansion with attention) model is a modified CNN architecture \cite{yu2016deep}.
LACE, first proposed in \cite{yu2016deep} and depicted in  Figure~\ref{fig:LACE},
is a variant of time-delay neural network (TDNN) \cite{waibel1989phoneme} in which each higher layer is a weighted sum of
nonlinear transformations of a window of lower layer frames.
Lower layers focus on extracting simple local patterns while higher layers extract complex patterns that cover broader contexts.
Since not all frames in a window carry the same importance, a learned attention mask is applied, shown as the ``element-wise matrix product''
in Figure~\ref{fig:LACE}.
The LACE model thus differs from the earlier TDNN models \cite{waibel1989phoneme,waibel1989consonant} in this attention masking,
as well as the ResNet-like linear pass-through connections.
As shown in the diagram, the model is composed of four blocks, each with the same architecture.
Each block starts with a convolution layer with stride two, which sub-samples the input and increases the number of channels.
This layer is followed by four ReLU convolution layers with jump-links similar to those used in ResNet. 
As for ResNet, batch normalization \cite{batchnorm} is used between layers.

\subsection{Bidirectional LSTM}
	\label{sec:blstm}

For our LSTM-based acoustic models we use a
bidirectional architecture (BLSTM) \cite{graves2005framewise} without frame-skipping
\cite{sak2015fast}. The core model structure is the LSTM defined in 
\cite{sak2014long}. We found that using networks with more than six
layers did not improve the word error rate on the development set,
and chose 512 hidden units, per direction, per layer; this gave a 
reasonable trade-off between training time and final model accuracy.

BLSTM performance was significantly enhanced using a spatial smoothing technique,
first described in \cite{parity-techreport}.
Briefly, a two-dimensional topology is imposed on each layer, and
activation patterns in which neighboring units are correlated are rewarded.

\subsection{CNN-BLSTM}

\begin{table*}
\centering
\caption{Comparison of CNN layer structures and parameters}
	\label{tab:CNNs}
\begin{tabular}{|l|c|c|c|}
\hline
                        &  ResNet                                                                                       & LACE                                                                           & CNN-BLSTM                                \\
\hline
\hline
Number of parameters    &  38M                                                                                                & 65M                                                                            & 48M                                      \\
\hline
Number of weight layers &  49                                                                                                 & 22                                                                             & 10                                       \\
\hline
Input                   &  40x41                                                                                              & 40x61                                                                          & 40x7xt                                   \\
\hline
Convolution\_1          &  \begin{tabular}[c]{@{}l@{}}{[}conv 1x1, 64\\  conv 3x3, 64\\  conv 1x1, 256{]} x 3\end{tabular}    & jump block {[}conv 3x3, 128{]} x 5                                             & \begin{tabular}[c]{@{}l@{}}{[}conv 3x3, 32, \\ padding in feature dim.{]} x 3 \end{tabular} \\
\hline
Convolution\_2          &  \begin{tabular}[c]{@{}l@{}}{[}conv 1x1, 128\\  conv 3x3, 128\\  conv 1x1, 512{]} x 4\end{tabular}  & jump block {[}conv 3x3, 256{]} x 5                                             &                                          \\
\hline
Convolution\_3          &  \begin{tabular}[c]{@{}l@{}}{[}conv 1x1, 256\\  conv 3x3, 256\\  conv 1x1, 1024{]} x 6\end{tabular} & jump block {[}conv 3x3, 512{]} x 5                                             &                                          \\
\hline
Convolution\_4          &  \begin{tabular}[c]{@{}l@{}}{[}conv 1x1, 512\\  conv 3x3, 512\\  conv 1x1, 2048{]} x 3\end{tabular} & jump block {[}conv 3x3, 1024{]} x 5                                            &                                          \\
\hline
BLSTM                   &                                                                                                     &                                                                                & {[} blstm, cells = 512{]} x 6            \\
\hline
Output                  &  \begin{tabular}[c]{@{}l@{}}average pool\\ Softmax (9k or 27k)\end{tabular}                              & \begin{tabular}[c]{@{}l@{}}{[}conv 3x4, 1{]} x 1\\ Softmax (9k or 27k)\end{tabular} & Softmax (9k or 27k) \\
\hline
\end{tabular}
\end{table*}

A new addition to our system this year is a CNN-BLSTM model inspired by \cite{SainathEtAl:icassp2015}.
Unlike the original BLSTM model, we included the context of each time point as an input feature in the model.
The context windows was $[-3, 3]$, so the input feature has size 40x7x$t$,
with zero-padding in the frequency dimension, but not in the time dimension.
We first apply three convolutional layers on the features at time $t$, and then apply six BLSTM layers to the resulting time sequence,
similar to structure of our pure BLSTM model. 

Table~\ref{tab:CNNs} compares the layer structure and parameters of the two pure CNN architectures, as well as the CNN-BLSTM.

\subsection{Senone set diversity}

One standard element of state-of-the-art ASR systems is the combination of multiple acoustic models.
Assuming these models are {\em diverse}, i.e., make errors that are not perfectly correlated,
an averaging or voting combination of these models should reduce error.
In the past we have relied mainly on different model architectures to produce diverse acoustic models.
However, results in \cite{parity-techreport} for multiple BLSTM models showed that 
diversity can also be achieved using different sets of senones (clustered subphonetic units).
Therefore, we have now adopted a variety of senone sets for all model architectures.
Senone sets differ by clustering detail (9k versus 27k senones), as well as two slightly different phone sets and corresponding dictionaries.
The standard version is based on the CMU dictionary and phone set (without stress, but including a schwa phone).
An alternate dictionary adds specialized vowel and nasal phones used exclusively for filled pauses and backchannel words, inspired by \cite{sri-2000}.
Combined with set sizes, this gives us a total of four distinct senone sets.

\subsection{Speaker adaptation}

Speaker adaptive modeling in our system is based on 
conditioning the network on an i-vector \cite{dehak2011front}  
characterization of each speaker \cite{saon2013speaker,saonSRK16}.
A 100-dimensional i-vector is generated for each conversation side
(channel A or B of the audio file, i.e., all the speech coming from the same speaker).
For the BLSTM systems,
the conversation-side i-vector $v_s$ is appended to each frame of input.
For convolutional networks, this approach is inappropriate because
we do not expect to see spatially contiguous patterns in the input.
Instead, for the CNNs, we add a learnable weight matrix $W^l$ to each 
layer, and add $W^l v_s$ to the activation of the layer before the 
nonlinearity. Thus, in the CNN, the i-vector essentially serves as an 
speaker-dependent bias to each layer.

For results showing the effectiveness of i-vector adaptation on our models,
see \cite{ms-swb-icassp2017}.

\subsection{Sequence training}

All our models are sequence-trained using maximum mutual information (MMI)
as the discriminative objective function.
Based on the approaches of \cite{chen2006advances} and \cite{povey2016purely},
the denominator graph is a full trigram LM over phones and senones.
The forward-backward computations are cast as matrix operations, and 
can therefore be carried out efficiently on GPUs without requiring a
lattice approximation of the search space.
For details of our implementation and empirical evaluation relative
to cross-entropy trained models, see \cite{ms-swb-icassp2017}.

\subsection{Frame-level model combination}

In our new system we added frame-level combination of senone posteriors from
multiple acoustic models.
Such a combination of neural acoustic models is effectively just another, albeit more complex,
neural model.
Frame-level model combination is constrained by the fact that the underlying senone sets 
must be identical.

\begin{table*}
\centering
\caption{Acoustic model performance by senone set, model architecture, and for various frame-level combinations, using an N-gram LM.
The ``puhpum'' senone sets use an alternate dictionary with special phones for filled pauses.}
	\label{tab:acoustic}

\begin{tabular}{|l|l|c|c|}
\hline
Senone set	& Architecture			& devset WER	& test WER 	\\	
\hline
\bf 9k		& BLSTM				& 11.5		& 8.3		\\	
		& ResNet			& 10.0		& 8.2		\\	
		& LACE				& 11.2		& 8.1		\\	
		& CNN-BLSTM			& 11.3		& 8.4		\\	
		& BLSTM+ResNet+LACE		& 9.8		& 7.2		\\	
		& BLSTM+ResNet+LACE+CNN-BLSTM	& 9.6		& 7.2		\\	
\hline
\bf 9k puhpum	& BLSTM				& 11.3		& 8.1		\\	
		& ResNet			& 11.2		& 8.4		\\	
		& LACE				& 11.1		& 8.3		\\	
		& CNN-BLSTM			& 11.6		& 8.4		\\	
		& BLSTM+ResNet+LACE		& 9.7		& 7.4		\\	
		& BLSTM+ResNet+LACE+CNN-BLSTM	& 9.7		& 7.3		\\	
\hline
\bf 27k		& BLSTM				& 11.4		& 8.0		\\	
		& ResNet			& 11.5		& 8.8		\\	
		& LACE				& 11.3		& 8.8		\\	
		& BLSTM+ResNet+LACE		& 10.0		& 7.5		\\	
\hline
\bf 27k puhpum	& BLSTM				& 11.3		& 8.0		\\	
		& ResNet			& 11.2		& 8.0		\\	
		& LACE				& 11.0		& 8.4		\\	
		& BLSTM+ResNet+LACE		& 9.8		& 7.3		\\	
\hline
\end{tabular}
\end{table*}

Table~\ref{tab:acoustic} shows the error rates achieved by various 
senone set, model architectures, and frame-level combination of multiple architectures.
The results are based on N-gram language models, and all combinations are equal-weighted.

\section{Language Models}
	\label{sec:language}

\subsection{Vocabulary size}

In the past we had used a relatively small vocabulary of 30,500 words drawn only from
in-domain (Switchboard and Fisher corpus) training data.
While this yields an out-of-vocabulary (OOV) rate well below 1\%, our error rates have reached levels 
where even small absolute reductions in OOVs could potentially have a significant impact
on overall accuracy.
We supplemented the in-domain vocabulary with the most frequent words in the out-of-domain
sources also used for language model training:  the LDC Broadcast News corpus and the UW Conversational Web corpus.
Boosting the vocabulary size to 165k reduced the OOV rate (excluding word fragments) on the eval2002 devset from 0.29\% to 0.06\%.
Devset error rate (using the 9k-senones BLSTM+ResNet+LACE acoustic models, see Table~\ref{tab:acoustic}) dropped from 9.90\% to 9.78\%.

\subsection{LSTM-LM rescoring}

For each acoustic model our system decodes with a slightly pruned 4-gram LM and generates lattices.
These are then rescored with the full 4-gram LM to generate 500-best lists.
The N-best lists in turn are then rescored with LSTM-LMs.

Following promising results by other researchers \cite{SundermeyerEtAl:ieee2015,medennikov2016improving},
we had already adopted LSTM-LMs in our previous system, with a few enhancements \cite{parity-techreport}:
\begin{itemize}
\item
	Interpolation of models based on one-hot word encodings (with embedding layer)
	and another model using letter-trigram word encoding (without extra embedding layer).
\item
	Log-linear combination of forward- and backward-running models.
\item
	Pretraining on the large out-of-domain UW Web corpus (without learning rate adjustment),
	followed by final training on in-domain data only, with learning rate adjustment schedule.
\item
	Improved convergence through a variation of self-stabilization \cite{ghahremani2016stab}, in which each output vector $x$ of non-linearities are scaled by
	$\frac{1}{4}\ln(1 + e^{4\beta})$, where a $\beta$ is a scalar that is learned for each output.
	This has a similar effect as the scale of the well-known batch normalization technique \cite{batchnorm}, but can be used in recurrent loops.
\item
	Data-driven learning of the penalty to assign to words that occur in the decoder LM but not in the LSTM-LM vocabulary.
	The latter consists of all words occurring twice or more in the in-domain data (38k words).
\end{itemize}
Also, for word-encoded LSTM-LMs, we use the approach from \cite{press2016using} to tie the input embedding and output embedding together.

In our updated system, we add the following additional utterance-scoped LSTM-LM variants:
\begin{itemize}
\item
	A character-based LSTM-LM
\item
	A letter-trigram word-based LSTM-LM using a variant version of text normalization
\item
	A letter-trigram word-based LSTM-LM using a subset of the full in-domain training corpus (a result of holding out a portion of training data for perplexity tuning)
\end{itemize}
All LSTM-LMs with word-level input use three 1000-dimensional hidden layers.
The word embedding layer for the word-based is also of size 1000, and the letter-trigram encoding has size 7190 (the number of unique trigrams).
The character-level LSTM-LM uses two 1000-dimensional hidden layers, on top of a 300-dimensional embedding layer.

As before, we build forward and backward running versions of these models, and combine them additively in the log-probability space, using equal weights.
Unlike before, we combine the different LSTM-architectures via log-linear combination in the rescoring stage, rather than via linear interpolation at the word level.
The new approach is more convenient when the relative weighting of a large number of models needs to be optimized, and the optimization happens jointly with the
other knowledge sources, such as the acoustic and pronunciation model scores.

\begin{table}
\centering
\caption{Perplexities of utterance-scoped LSTM-LMs}
	\label{tab:lstm-lm-results}

\begin{tabular}{|l|c|c|c|}
\hline
Model structure				& Direction	& PPL		& PPL \\
					& 		& devset	& test	\\
\hline
Word input, one-hot 			& forward	& 50.95		& 44.69		\\
					& backward	& 51.08		& 44.72		\\
\hline
Word input, letter-trigram		& forward	& 50.76		& 44.55		\\
					& backward	& 50.99		& 44.76		\\
\hline
\quad + alternate text norm		& forward	& 52.08		& 43.87		\\
					& backward	& 52.02		& 44.23		\\
\hline
\quad + alternate training set		& forward	& 50.93		& 43.96		\\ 
					& backward	& 50.72		& 44.36		\\
\hline
Character input				& forward	& 51.66		& 44.24		\\
					& backward	& 51.92		& 45.00		\\
\hline
\end{tabular}
\end{table}

Table~\ref{tab:lstm-lm-results} shows perplexities of the various LSTM language models on dev and test sets.
The forward and backward versions have very similar perplexities, justifying tying their weights in the eventual score weighting.
There are differences between the various input encodings, but they are small, on the order of 2-4\% relative.

\subsection{Dialog session-based modeling}

The task at hand is not just to recognize isolated utterances but entire conversations.
We are already exploiting global conversation-level consistency via speaker adaptation,
by extracting i-vector from all the speech on one side of the conversation, as described earlier.
It stands to reason that the language model could also benefit from information beyond the 
current utterance, in two ways:  first conversations exhibit global coherence, especially in terms of 
conversation topic, and especially since Switchboard conversations are nominally on a pre-defined topic.
Lexical entrainment \cite{BrennanClark:jep96}, the tendency of conversants to adopt the same words and phrases,
could also be exploited for language modeling.
There is a large body of work on adaptation of language models to topics and otherwise \cite{Bellegarda:specom2004},
and recurrent neural network model extensions have been proposed to condition on global context
\cite{mikolov2012context}.
A second set of conversation-level phenomena operate beyond the utterance, but more locally.
Linguistic conversation analysis has long noted that utterance types come in {\em adjacency pairs} \cite{Schegloff:68},
with preferences for certain pairs over others (like a statement followed by agreement versus disagreement).
Therefore, words in an utterance should be more predicable based on the previous utterance,
as well as information about whether a speaker change occurred, and this has been proposed as useful information in language models \cite{JiBilmes:hlt2004}.
We can also include in the history information on whether utterances overlap, since overlap is partially predicted by the words spoken
\cite{ShribergEtAl:eurospeech2001}.
This type of conditioning information could help model dialog 
behaviors such as floor grabbing, back-channeling, and collaborative completions.

\begin{figure}
\centering
\includegraphics[width=.5\textwidth]{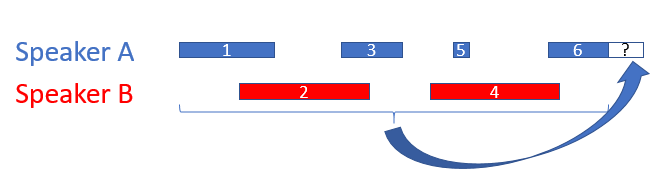}
\caption{Use of conversation-level context in session-based LM}
	\label{fig:session-lm}
\end{figure}

In order to capture both global and local context for modeling the current utterance, we train session-based
LSTM-LMs.
We serialize the utterances in a conversation based on their onset times (using the waveform cut points as approximate
utterance onset and end times).
We then string the words from both speakers together to predict the following word at each position, as depicted in 
Figure~\ref{fig:session-lm}.
Optionally, extra bits in the input are used to encode whether a speaker change occurred, or whether the current utterance overlaps
in time with the previous one.
When evaluating the session-based LMs on speech test data, we use the 1-best hypotheses from the N-best generation step (which uses only
an N-gram LM) as a stand-in for the conversation history.

Table~\ref{tab:session-lm-results} shows the effect of session-level modeling and of these optional elements on the model perplexity.
There is a large perplexity reduction of 21\% by conditioning on the previous word context, with smaller incremental reductions 
from adding speaker change and overlap information.
The table also compares the word error rate with the full session-based model to the baseline, within-utterance LSTM-LM.
As shown in the last row of the table, some of the perplexity gain over the baseline is negated by the use of 1-best recognition output
for the conversation history.
However, the perplexity degrades by only 7-8\% relative due to the noisy history.

\begin{table}
\centering
\caption{Perplexities and word errors with session-based LSTM-LMs (forward direction only).
The last line reflects the use of 1-best recognition output for words in preceding utterances.}
	\label{tab:session-lm-results}

\begin{tabular}{|l|c|c|c|c|}
\hline
Model inputs				& PPL		& PPL 	& WER		& WER \\
					& devset	& test	& devset	& test \\
\hline
Utterance words, letter-3grams		& 50.76		& 44.55 & 9.5 	& 6.8		\\
 + session history words		& 39.69		& 36.95	&	& 		\\
\quad + speaker change			& 38.20		& 35.48	&	&		\\
\quad \quad + speaker overlap 		& 37.86		& 35.02	&  	& 		\\
 \quad \quad (with 1-best history)	& 40.60		& 37.90 & 9.3	& 6.7		\\
\hline
\end{tabular}
\end{table}

For inclusion in the overall system, we built letter-trigram and word-based versions of the session-based LSTM (in both directions).

Both session-based LM scores are added to the utterance-based LSTM-LMs described earlier for log-linear combination.

\section{Experimental Setup}

\subsection{Data}

The data sets used for system training are unchanged \cite{parity-techreport};
they consist of the public and shared data sets used in the 
DARPA research community.
Acoustic training used the English CTS (Switchboard and Fisher) corpora.
Language model training, in addition, used the English CallHome transcripts, the 
BBN Switchboard-2 transcripts, the LDC Hub4 (Broadcast News) corpus, and the UW conversational web corpus \cite{BulykoEtAl:hlt2003}.
Evaluation is carried out on the NIST 2000 CTS test set Switchboard portion.
The Switchboard-1 and Switchboard-2 portions of the NIST 2002 CTS test set were used for tuning and development.

\subsection{Model training}

All neural networks in the final system were trained with the
Microsoft Cognitive Toolkit, or CNTK \cite{CNTK,cntkai} on a Linux-based multi-GPU server farm.
CNTK allows for flexible model definition, while at the same
time scaling very efficiently
across multiple GPUs {\em and} multiple servers.
The resulting fast experimental turnaround
using the full 2000-hour corpus was critical for our work.

Training the acoustic models in this paper on a single GPU would take many weeks or even months.
CNTK made training times feasible by parallelizing the stochastic gradient descent (SGD) training
with a {\em 1-bit SGD}
parallelization technique \cite{seide20141}. This data-parallel method distributes minibatches over multiple worker nodes, and then aggregates the sub-gradients.
While the necessary communication time would otherwise be prohibitive,
the 1-bit SGD method eliminates the bottleneck by combining 1-bit quantization of gradients and automatic minibatch-size scaling,
as described in more detail in \cite{parity-techreport}.

We use the CNTK ``FsAdaGrad'' learning algorithm, which is an implementation of Adam \cite{Adam}.
A typical learning rate is $3\times 10^{-6}$,  
and learning rates are automatically adjusted with a decrease factor of 0.7.
Momentum is set at a constant value of 2500 throughout model training.
For individual acoustic models, we find that training converges after 1.5 to 2 passes over the 2000-hour training set.
We do not use dropout or gradient noise in our model training,
only the aforementioned spatial smoothing technique for BLSTM model training.

\section{System Combination and Results}
	\label{sec:combination}

\subsection{Confusion network combination}

After rescoring all system outputs with all language models, we 
combine all scores log-linearly and normalize to estimate utterance-level posterior probabilities.
All N-best outputs for the same utterance are then concatenated and 
merged into a single word confusion network (CN), using the SRILM nbest-rover tool \cite{stolcke2002srilm,sri-2000}.

\subsection{System Selection}

Unlike in our previous system, we do not apply estimated, system-level weights to the posterior probabilities estimated
from the N-best hypotheses.  All systems have equal weight upon combination.
This simplification allows us to perform a brute-force search over all possible subsets of systems,
picking the ones that give the lowest word error on the development set.
We started with 9 of our best individual systems, and eliminated two, leaving a combination of 7 systems,

\subsection{Confusion network rescoring and backchannel modeling}

As a final processing step, we generate new N-best lists from the confusion networks resulting from system combination.
Following \cite{BangaloreEtAl:asru2001}, these are once more rescored using the N-gram LM,
but also with a subset of the utterance-level LSTM-LMs, and one additional knowledge source.
The word log posteriors from the confusion network take the place of the acoustic model scores in this final rescoring step.

The additional knowledge source at this stage was motivated by our analysis of differences between machine versus human
transcription errors \cite{StolckeDroppo:interspeech2017}.
We found that the major machine-specific error pattern is a misrecognition of filled pauses (`uh', `um') as backchannel acknowledgments ('uh-huh', `mhm'). 
In order to allow the system learn a correction for this problem, we provide the number of backchannel tokens in a hypotheses as a pseudo-score and allow the 
score weight optimization to find a penalty for it. (Indeed, a negative weight is learned for the backchannel count.)

Table~\ref{tab:final-results} compares the individual systems that were selected for combination, before and after rescoring with LSTM-LMs,
and then shows the progression of results in the final processing stages,
starting with the LM-rescored individual systems, the system combination, and the CN rescoring.
The collection of LSTM-LMs (which includes the session-based LMs) gives a very consistent 22 to 25\% relative error reduction on individual systems,
compared to the N-gram LM.
The system combination reduces error by 4\% relative over the best individual systems, and the CN rescoring improves another 2-3\% relative.

\begin{table*}
\centering
\caption{Results for LSTM-LM rescoring on systems selected for combination, the combined system, and confusion network rescoring}
	\label{tab:final-results}

\begin{tabular}{|l|l|c|c|c|c|}
\hline
Senone set	& Model/combination step	& WER		& WER		& WER		& WER \\
		&				& devset	& test		& devset	& test	\\
		&				& \multicolumn{2}{c|}{ngram-LM} & \multicolumn{2}{c|}{LSTM-LMs} \\
\hline
9k		& BLSTM				& 11.5		& 8.3		& 9.2		& 6.3	\\	
27k		& BLSTM 			& 11.4		& 8.0		& 9.3		& 6.3	\\	
27k-puhpum	& BLSTM 			& 11.3		& 8.0		& 9.2		& 6.3 	\\	
9k 		& BLSTM+ResNet+LACE+CNN-BLSTM	& 9.6		& 7.2		& 7.7		& 5.4	\\	
9k-puhpum	& BLSTM+ResNet+LACE		& 9.7		& 7.4		& 7.8		& 5.4	\\	
9k-puhpum	& BLSTM+ResNet+LACE+CNN-BLSTM	& 9.7		& 7.3		& 7.8		& 5.5 	\\	
27k		& BLSTM+ResNet+LACE		& 10.0		& 7.5		& 8.0		& 5.8	\\	
\hline
-		& Confusion network combination		&	&		& 7.4		& 5.2	\\
-		& + LSTM rescoring			&	&		& 7.3		& 5.2	\\
-		& \quad + ngram rescoring		&	&		& 7.2		& 5.2	\\
-		& \quad \quad + backchannel penalty	&	&		& 7.2		& 5.1	\\
\hline
\end{tabular}
\end{table*}

\section{Conclusions and Future Work}
	\label{sec:concl}

We have described the latest iteration of our conversational speech recognition system.
The acoustic model was enhanced by adding a CNN-BLSTM system, and the more systematic use of a variety of senone sets,
to benefit later system combination.
We also switched to combining different model architectures first at the senone/frame level, resulting in several 
acoustic combined systems that are then fed into the confusion-network-based combination at the word level.
The language model was updated with larger vocabulary (lowering the OOV rate by about 0.2\% absolute),
additional LSTM-LM variants for rescoring, and most importantly, session-level LSTM-LM that can model global and local coherence 
between utterances, as well as dialog phenomena.
The session-level model gives over 20\% relative perplexity reduction.
Finally, we introduce a confusion network rescoring step with special treatment for backchannels (based on a prior error analysis),
that gives a small additional gain after systems are combined.
Overall, we have reduced error rate for the Switchboard tasks by 12\% relative, from 5.8\% for the 2016 system, to now 5.1\%.
We note that this level of error is on par with the multi-transcriber error rate previously reported on the same task.

Future work we plan from here includes a more thorough evaluation, including on the CallHome genre of speech.
We also want to gain a better understanding of the linguistic phenomena captured by the session-level language model,
and reexamine the differences between human transcriber and machine errors.

{\bf Acknowledgments.}
We wish to thank our colleagues Hakan Erdogan, Xiaodong He, Jinyu Li, Frank Seide, Mike Seltzer, and Takuya Yoshioka for their
valued input during system development, and ICSI for assistance with CTS data sets.

\bibliographystyle{ieee-shortnames}
\bibliography{strings,refs}

\end{document}